\documentclass[letterpaper, 10 pt, conference]{ieeeconf}  

\IEEEoverridecommandlockouts
\usepackage{cite}
\usepackage{amsmath,amssymb,amsfonts}
\usepackage{algorithmic}
\usepackage{graphicx}
\usepackage{textcomp}
\usepackage{xcolor}
\usepackage{colortbl}
\usepackage{array}
\usepackage{algorithm}
\usepackage{booktabs}
\usepackage{multirow}
\usepackage{url}
\usepackage{balance}
\usepackage{microtype}
\usepackage[T1]{fontenc}
\usepackage{pifont}

\newcommand{\cmark}{\ding{51}}
\newcommand{\xmark}{\ding{55}}

\def\BibTeX{{\rm B\kern-.05em{\sc i\kern-.025em b}\kern-.08em
    T\kern-.1667em\lower.7ex\hbox{E}\kern-.125emX}}

\title{\LARGE \bf
TIDAL: Temporally Interleaved Diffusion and Action Loop for Dynamic Manipulation
}

\author{
\begin{tabular}{c}
Yuteng Sun$^{1,2}$, Haoran Wang$^{1,3}$, Ruofei Bai$^{1,3}$, Zhengguo Li$^{1}$, Jun Li$^{1}$, \\
Wenyu Liang$^{1}$, Meng Yee (Michael) Chuah$^{1}$, and Wei-Yun Yau$^{1,\ast}$
\end{tabular}%
\thanks{$^{1}$Authors are with the Institute of Advanced Intelligence and Computing (IAIC), Agency for Science, Technology and Research (A*STAR), Singapore.}%
\thanks{$^{2}$Tsinghua University, Beijing, China.}%
\thanks{$^{3}$Nanyang Technological University, Singapore.}%
\thanks{$^{\ast}$Correspondence: {\tt\small Yau\_Wei\_Yun@a-star.edu.sg}.}%
}

\begin{document}

\maketitle
\thispagestyle{empty}
\pagestyle{empty}

\begin{abstract}
Large-scale Vision-Language-Action (VLA) models offer semantic generalization but suffer from high inference latency because they adopt a low-frequency batch-and-execute paradigm. This frequency mismatch creates an execution blind spot, causing failures in dynamic environments where targets move during the open-loop execution window. We propose TIDAL (Temporally Interleaved Diffusion and Action Loop), a hierarchical framework that decouples semantic reasoning from high-frequency actuation. TIDAL operates as a backbone-agnostic scheduler for diffusion-based VLAs, using a dual-frequency architecture to redistribute the computational budget. Specifically, a low-frequency macro-intent loop caches semantic embeddings, while a high-frequency micro-control loop interleaves single-step flow integration with execution, conditioning on the latest state fused with motion cues. To handle the resulting latency shift, we introduce a temporally misaligned training strategy where the policy learns staleness-aware compensation, conditioning on stale semantic intent alongside real-time proprioception. TIDAL is architectural, making it orthogonal to system-level optimizations. Experiments show an average 2.5x performance gain over open-loop baselines in dynamic interception tasks. Despite a slight decrease in static success rates, our approach yields a 4x increase in feedback frequency and extends the effective horizon of semantic embeddings beyond the native action chunk size. Under non-paused physics and on a real robot, TIDAL remains robust to unmodeled dynamics, while standard open-loop baselines fail due to latency-induced error accumulation.
\end{abstract}

\section{Introduction}
\begin{figure*}[t]
\centering
\includegraphics[width=\linewidth]{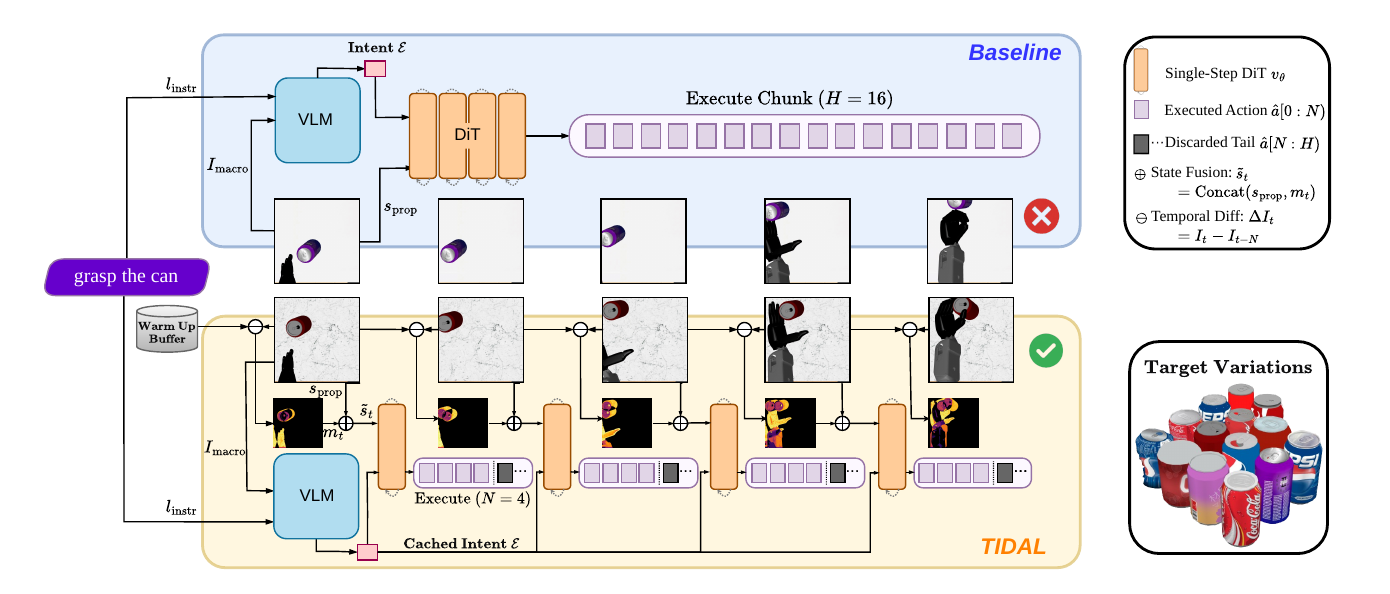}
\caption{Comparison of inference paradigms for dynamic interception under target morphology variation given the instruction ``grasp the can.''
\textit{(Top) Standard Baseline:} The system performs blocking inference at the start and generates a fixed 16-step action chunk conditioned on the initial macro observation $I_{\text{macro}}$ and proprioception $s_{\text{prop}}$ via intent $\mathcal{E}$. Open-loop execution follows a stale trajectory and can miss the moving target.
\textit{(Bottom) TIDAL (Ours):} The macro-loop queries the VLM once to cache intent $\mathcal{E}$, while the micro-loop performs single-step interleaved flow integration using the latest fused state $\tilde{s}*t$ from proprioception $s*{\text{prop}}$ and motion cues $m_t$. The policy executes only the immediate chunk ($N=4$) and discards the tail, refreshing control at $\sim$9,Hz for robust interception.}
\label{fig:main_arch}
\end{figure*}

The pursuit of generalist embodied agents has been accelerated by Vision-Language-Action models, exemplified by systems like GR00T \cite{bjorck2025gr00t} and $\pi_0$ \cite{black2024pi_0}. These models increasingly adopt hierarchical architectures that integrate a large Vision-Language Model with continuous diffusion or flow-matching heads. While these models offer strong semantic grounding from large-scale pre-training, their scale creates a severe computational bottleneck. Current VLAs typically operate on a batch-and-execute paradigm: the system pauses to process observations and computes an entire long-horizon action chunk, which is then executed open-loop. This introduces a fundamental frequency mismatch. Dynamic manipulation requires high-frequency control, whereas large-scale VLA inference often incurs substantial latency and struggles to satisfy such control demands in practice \cite{11164279, shao2025large}. In real-world deployment, this bottleneck is worsened by the mandatory execution time of the generated action chunk \cite{black2025realtime, tang2025vlash}. The combination of inference latency and open-loop execution creates an execution blind spot, where the robot remains unresponsive to environmental changes, often leading to failures. Simply increasing the replanning frequency is often compute-prohibitive. Higher replanning rates can increase end-to-end delay, making newly generated trajectories stale by the time they are executed under non-paused physics, where the environment continues to evolve during policy inference rather than waiting for computation to finish. On edge hardware with a fixed compute budget, increasing the control update rate is better achieved by scheduling rather than by more frequent backbone queries.

Existing latency mitigation strategies incur significant trade-offs. System-level optimizations, such as quantization (e.g., BitVLA \cite{wang2025bitvla}) or token pruning (e.g., FlashVLA \cite{tan2025think}), provide speedups but remain bounded by the sequential attention mechanism. Distilling small-scale policies like TinyVLA \cite{10900471} trades rich semantic priors for throughput. More recently, asynchronous scheduling \cite{black2025realtime} and dual-system architectures \cite{chen2025fastinslow} have attempted to decouple execution from reasoning. However, these methods often require careful stream management or separate policy networks. We argue that high-frequency control should not require losing the semantic intelligence of large foundation models, nor should it require training distinct policies.

We propose TIDAL (Temporally Interleaved Diffusion and Action Loop), a backbone-agnostic framework that bridges this frequency gap via compute-efficient interleaved scheduling. Our core insight is that semantic intent exhibits high temporal persistence, whereas physical state is highly transient. Consequently, the expensive VLM backbone need not be queried at every step. Instead of discarding the semantic context, we extend its lifespan by fusing it with high-frequency, real-time proprioceptive updates. TIDAL employs a dual-frequency scheduling strategy: a low-frequency macro-loop caches semantic embeddings, while a high-frequency micro-loop interleaves single-step flow integration with execution. This redistributes the computational budget, refreshing the action chunk every $\sim$110\,ms ($\approx 9$\,Hz) compared to the standard $\sim$400\,ms baseline cycle, yielding a $4\times$ increase in feedback frequency. Although this imposes a higher optimization burden that slightly impacts static task convergence, it provides the critical reactivity required for dynamic interaction and extends the lifespan of semantic embeddings beyond the native action chunk size.

Deploying this decoupled architecture introduces a temporal misalignment between the stale semantic condition and the current physical state. To address this, we introduce a temporally misaligned training strategy where the policy learns to compensate for intent staleness, using current proprioception signals to reconcile stale semantic conditions with the rapidly evolving physical state. Additionally, we resolve the velocity insensitivity of static vision encoders by incorporating a Differential Motion Predictor with contact-gated injection.

Our primary contributions are:
\begin{enumerate}
    \item A serial, time-interleaved scheduler for VLAs that increases control update rate without increasing backbone calls.
    \item A training paradigm that enables robust compensation for stale semantic cues under temporal misalignment.
    \item A flow-matching inference/training recipe that enables reliable one-step Euler chunk generation for high-frequency interleaving.
    \item TIDAL achieves an average 2.5$\times$ performance gain in dynamic interception tasks. Robust performance under non-paused inference evaluation and real-world deployment on a physical manipulator validates its efficacy.
\end{enumerate}

\section{Related Work}

\subsection{Latency Mitigation and Real-Time Inference}
Current methods bridge the inference-actuation gap via model simplification, adaptive computation, or parallel scheduling.

\subsubsection{Model Distillation and Architectural Simplification}
One strategy improves deployment efficiency through lightweight VLA variants, compressed backbones, or efficient architectures, including compact models \cite{shukor2025smolvla,hung2025nora,10900471} and state-space architectures such as \cite{NEURIPS2024_46a12649}. These methods typically trade model capacity or compatibility for speed, potentially weakening the semantic reasoning of larger backbones. TIDAL instead retains the full backbone and improves responsiveness through scheduling.

\subsubsection{Static Compression and Adaptive Inference}
Static compression techniques like quantization \cite{wang2025bitvla} and token pruning \cite{tan2025think, yang2025efficientvla} directly reduce memory footprint and computational cost. Dynamic inference mechanisms further accelerate forward passes via confidence-based early exits \cite{song2025ceed}, layer routing \cite{zhang2025mole}, and parallel decoding \cite{11247519}. Alongside system-level kernel fusion \cite{ma2025running}, these methods reduce per-step computation but remain bound by the sequential observe-then-act bottleneck. TIDAL is orthogonal to these optimizations; it acts as a temporal scheduler that can leverage such accelerated backbones to further densify replanning cycles.

\subsubsection{Asynchronous Pipeline and Future Awareness}
Recent works optimize the inference pipeline via asynchronous execution to break sequential dependencies. Real-Time Chunking \cite{black2025realtime} inpaints future actions during execution, while VLASH \cite{tang2025vlash} uses a future-state-aware paradigm. VLASH trains the policy to compensate for latency by conditioning on predicted future states. Similarly, Spec-VLA \cite{wang-etal-2025-spec} accelerates decoding via speculative sampling. These approaches reduce latency exposure by overlapping computation with execution or by forecasting future states. In contrast, under a serial compute constraint and non-paused physics, reactive updates from the latest state become critical.
TIDAL shares the insight of training with temporal misalignment but diverges in execution. Pipelined approaches require high VLM throughput to minimize the prediction horizon. In contrast, TIDAL's lightweight micro-loop reduces effective calculation time to minimal levels. This allows TIDAL to condition directly on the current fused state, avoiding the complexity and uncertainty of predicting future system states.

\subsection{Decoupled Architectures and Temporal Redundancy}

Physical states change rapidly, yet high-level semantic intent persists over time \cite{guan2025efficient}. While methods like VLA-Cache \cite{xu2025vlacache} exploit this by reusing visual features, recent frameworks like DuoCore-FS \cite{zou2025asynchronous} and Fast-in-Slow \cite{chen2025fastinslow} take this further by decoupling reasoning from motor control. They implement a parallel architecture: a slow VLM updates a semantic buffer, while a fast policy queries it at high frequency. However, running two models concurrently demands significant memory bandwidth and parallel compute, which are often prohibitive for edge devices.

To avoid parallel execution requirements, TIDAL achieves similar decoupling but shifts from a \textit{parallel} to an \textit{interleaved} strategy. By treating the VLM and action head as serial, temporally multiplexed components, we eliminate the need for parallel hardware. We freeze the heavy semantic embedding and use a single-step flow integration mechanism for the high-frequency micro-loop. This approach ensures real-time reactivity within a strict serial backbone compute budget, which we validate under realistic, non-paused physical constraints.

\subsection{Action Chunking and Execution Strategies}
Action chunking \cite{Zhao-RSS-23} distributes the computational cost of VLA backbones \cite{pmlr-v270-kim25c, Xu_2025_ICCV}. Recent works like Real-Time Chunking \cite{black2025realtime} improve throughput by overlapping inference with execution via inpainting or chunk splicing. However, these asynchronous methods condition actions on stale observations captured at the start of the inference delay.

TIDAL moves away from this rigid predict-then-execute paradigm. Leveraging Flow Matching \cite{liu2023flow}, we reinterpret chunk generation as repeated conditional vector-field queries rather than a single static open-loop rollout. By injecting the latest proprioceptive state into the action head at each micro-update, our interleaved integration enables high-frequency reactivity without the complexity of asynchronous stream management or full sequence regeneration.

\section{Methodology}

\subsection{Hierarchical Dual-Frequency Inference Scheduling}

TIDAL splits VLA inference into two nested processes: a low-frequency macro-intent loop for semantic reasoning and a high-frequency micro-control loop for closed-loop refinement. This decomposition enables high-rate control updates while keeping the 3B-parameter backbone at a low query rate by caching semantic embeddings and interleaving flow matching integration steps.

\subsubsection{Macro-Intent Loop (Semantic Caching)}
The macro-loop extracts high-level intent from visual observations. Let $l_{\text{instr}}$ denote the natural language instruction and $I_{t_m}$ denote the RGB observation at macro-step $m$. We use a pre-trained VLM backbone $\Phi_{\text{VLM}}$ to map these inputs to a dense intent embedding $\mathcal{E}_m$:
\begin{equation}
    \mathcal{E}_m = \Phi_{\text{VLM}}(I_{t_m}, l_{\text{instr}})
\end{equation}
The embedding $\mathcal{E}_m$ is stored in a latent intent buffer and remains frozen for the macro-cycle (horizon $H=16$). This caching strategy isolates the control loop from the high latency of $\Phi_{\text{VLM}}$, exploiting the fact that semantic intent persists over short execution horizons.

\subsubsection{Micro-Control Loop (Interleaved Flow Execution)}

The micro-loop functions as a reactive policy that queries the action head at high frequency. To align with the standard execution horizon ($H=16$) while maximizing feedback density, we structure the inference as a cascade of $K=4$ interactions per macro-cycle.
Within each macro-cycle, we re-index time locally so that the macro observation is at $t=0$. Given a cached intent $\mathcal{E}$, the loop iterates through latency stages $k \in \{0, \dots, K-1\}$. At each stage, the loop observes the current state $\tilde{s}_{k \cdot N}$. To reduce inference latency, we bypass multi-step ODE solvers and generate the full-horizon action trajectory $\mathbf{\hat{a}} \in \mathbb{R}^{H \times D_a}$ via a single-step Euler approximation ($\Delta \tau = 1$) from $\tau=0$ to $1$:
\begin{equation}
    \mathbf{\hat{a}} = \mathbf{x}_0 + \Delta\tau \, v_\theta(\mathbf{x}_0, 0, \tilde{s}_{k \cdot N}, \mathcal{E}), \qquad \Delta\tau = 1.
\end{equation}
where $\mathbf{x}_0 \sim \mathcal{N}(\mathbf{0}, \mathbf{I})$ is the source noise trajectory at $\tau=0$, and $v_\theta$ predicts the conditional flow-matching velocity field along the optimal-transport path. We treat this as a one-step flow map optimized to be accurate near the source frontier ($\tau\approx0$), rather than a generic ODE solver.
After chunk generation, we execute only the first $N$ steps ($\mathbf{\hat{a}}_{[0,N)}$) on the robot hardware, discarding the rest ($\mathbf{\hat{a}}_{[N,H)}$).

TIDAL distributes the typical integration budget across the execution timeline. Inspired by streaming execution paradigms \cite{jiang2025streaming, 11127858}, we use a single Euler step as a latency-efficient update for the interleaved control loop at the chosen chunk size. By querying this field at every chunk boundary using the latest state $\tilde{s}_{k \cdot N}$, TIDAL actively corrects action chunk drift without increasing the backbone compute budget.

\begin{algorithm}[t]
\caption{TIDAL Hierarchical Inference Logic}
\label{alg:tidal}
\begin{algorithmic}[1]
\REQUIRE VLM Backbone $\Phi_{\text{VLM}}$, Policy Vector Field $v_\theta$, Prediction Horizon $H=16$, Execution Chunk $N=4$, Latency Stages $K=4$, Micro-Cam Buffer $\mathcal{B}$ (single-frame, initialized by warm-up for $N$ steps)

\LOOP
    \STATE \COMMENT{\textbf{Macro-Intent Loop}}
    \STATE Capture high-res image $I_{\text{macro}}$
    \STATE $\mathcal{E} \leftarrow \Phi_{\text{VLM}}(I_{\text{macro}}, l_{\text{instr}})$ \COMMENT{Freeze semantic intent}
    \STATE \COMMENT{\textbf{Micro-Control Loop}}
    \FOR{$k = 0$ \TO $K-1$}
        \STATE Capture proprioception $s_{\text{prop}}$ and micro frame $I_{k \cdot N}$
        \STATE $I_{\mathcal{B}} \leftarrow \mathcal{B}$ \COMMENT{previous micro frame}
        \STATE $\mathcal{B} \leftarrow I_{k \cdot N}$ \COMMENT{update buffer}
        \STATE $m_{k \cdot N} \leftarrow \phi_{\text{motion}}\!\left(\mathcal{T}(I_{k \cdot N}) - \mathcal{T}(I_{\mathcal{B}})\right)$
        \STATE $\tilde{s}_{k \cdot N} \leftarrow \text{Fuse}(s_{\text{prop}}, m_{k \cdot N})$ \COMMENT{State Fusion}
        
        \STATE $\mathbf{x}_0 \sim \mathcal{N}(\mathbf{0}, \mathbf{I})$ \COMMENT{Sample fresh noise trajectory of length $H$}
        \STATE $\mathbf{\hat{v}} \leftarrow v_\theta(\mathbf{x}_0, \tau=0, \tilde{s}_{k \cdot N}, \mathcal{E})$ \COMMENT{Query vector field at source frontier}
        \STATE $\mathbf{\hat{a}} \leftarrow \mathbf{x}_0 + \mathbf{\hat{v}} \cdot 1$ \COMMENT{Single-step Euler integration ($\Delta \tau = 1$)}
        
        \STATE \textbf{Execute} $\mathbf{\hat{a}}_{[0,N)}$ \COMMENT{Run $N$ steps, discard $H-N$}
        \STATE \textbf{Wait} for execution 
    \ENDFOR
\ENDLOOP
\end{algorithmic}
\end{algorithm}

\subsection{Temporally Misaligned Training}

\begin{figure*}[t]
\centering
\includegraphics[width=\linewidth]{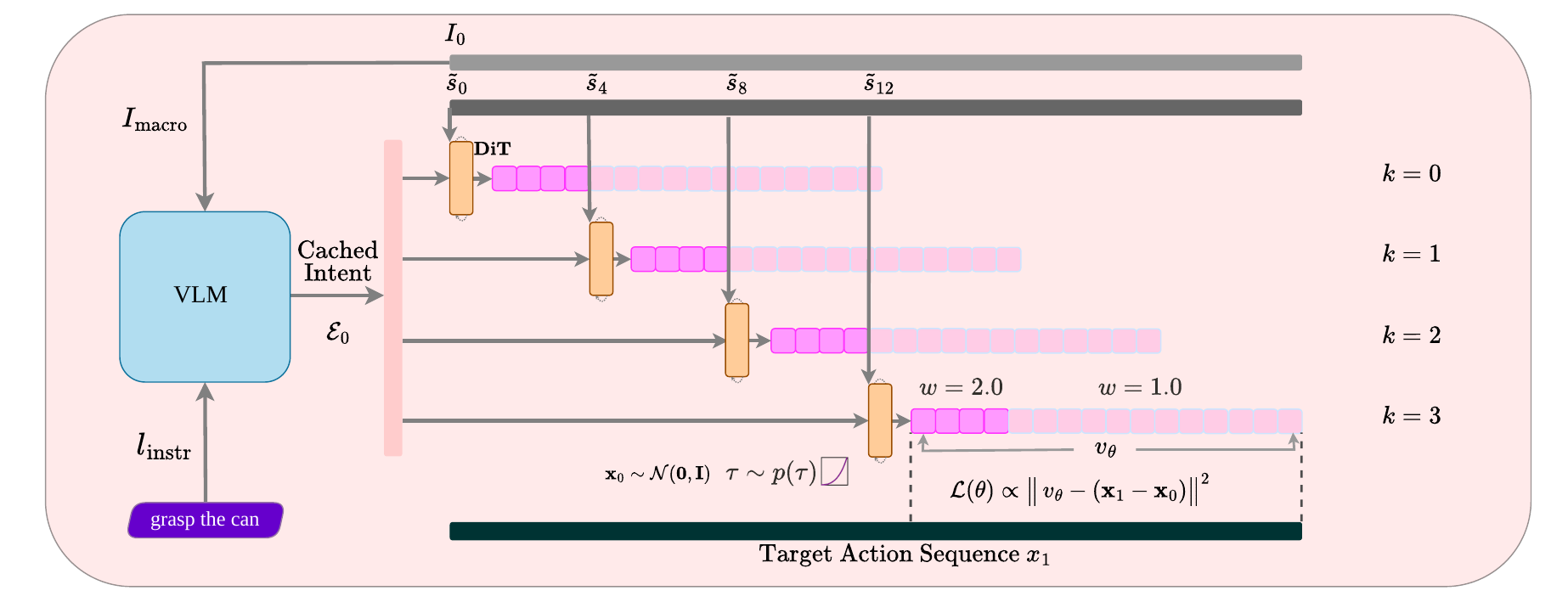}
\caption{\textit{Temporally Misaligned Training Strategy.} We decouple visual and physical timestamps to handle macro--micro latency.
\textit{Latency Injection:} A frozen semantic intent from $I_{\text{macro}}$ at $t{=}0$ is reused across latency stages ($k=0,\dots,3$) and paired with later fused states ($\tilde{s}*{0}, \tilde{s}*{4}, \tilde{s}*{8}, \tilde{s}*{12}$).
\textit{Optimization Objectives:} The DiT predicts the flow vector field $v_\theta$ over the $H{=}16$-step horizon, supervised by $(\mathbf{x}_1-\mathbf{x}_0)$. A horizon-weighted loss emphasizes the immediate execution prefix ($N{=}4$), while time-biased sampling concentrates $\tau$ near the noise source for single-step integration.}
\label{fig:training}
\end{figure*}

Synchronous training creates a distribution shift in our architecture because semantic conditions $\mathcal{E}$ ($I_{t-\Delta}$) lag behind the physical state $\tilde{s}_t$. Inspired by recent findings in asynchronous control \cite{tang2025vlash, zou2025asynchronous}, we address this via a temporally misaligned training strategy that explicitly decouples visual and physical input timestamps, enabling convergence despite semantic lag.

\subsubsection{Dynamic Latency Injection}
We simulate variable inference latencies by creating training samples with randomized time-lags. Instead of fixed windows $H$, we sample extended trajectory segments of length $L$:
\begin{equation}
    L = H + (K - 1) \times N
\end{equation}
where $H=16$ is the action horizon predicted by the policy, $N=4$ is the prefix executed before the next micro-update, and $K=4$ is the number of micro-updates performed under one cached macro intent. $L$ is the training segment length required so that every sampled latency stage still has an H-step supervision target. We set $L=28$.

During training, we discretize the potential delay into fixed stages, sampling a latency index $k \sim \mathcal{U}\{0, K-1\}$ for each iteration. The policy input tuple consists of:
\begin{itemize}
    \item Frozen Intent (at physical step $t=0$): Visual backbone input $I_0$.
    \item Current State (at physical step $t=k \cdot N$): Proprioceptive state $\tilde{s}_{k\cdot N}$.
    \item Action Target (at physical step $t=k \cdot N$): Ground truth sequence $\mathbf{a}_{[kN,\,kN+H)}$ (length $H$).
\end{itemize}
Restricting latency to a fixed set of discrete multiples of $N$, rather than sampling from a continuous interval, reduces the variance of the learning objective. This simplifies the optimization landscape, allowing the model to converge to distinct compensation modes for each stage of intent staleness, rather than fitting an infinite spectrum of delays.

\subsubsection{Horizon-Weighted \& Time-Biased Flow Matching}

The action chunk generator is implemented using Conditional Flow Matching (CFM) \cite{lipman2023flow}, which regresses a time-dependent vector field $v_\theta$. We define the optimal transport path $\psi_\tau(\mathbf{x}_0, \mathbf{x}_1) = (1-\tau)\mathbf{x}_0 + \tau \mathbf{x}_1$ between the Gaussian noise source $\mathbf{x}_0 \sim \mathcal{N}(\mathbf{0}, \mathbf{I})$ and the target action sequence $\mathbf{x}_1$. The objective minimizes the expected flow matching error:
\begin{equation}
\begin{split}
    \mathcal{L}(\theta) = \mathbb{E}_{k, \tau, \mathbf{x}_0, \mathbf{x}_1} \Bigg[ \sum_{i=0}^{H-1} w_i \cdot \Big\| & v_\theta(\psi_\tau(\mathbf{x}_0, \mathbf{x}_1), \tau, \tilde{s}_{k \cdot N}, \mathcal{E}_0)^{(i)} \\
    & - (\mathbf{x}_1^{(i)} - \mathbf{x}_0^{(i)}) \Big\|^2 \Bigg]
\end{split}
\end{equation}
where $\tau \in [0, 1]$ is the continuous flow time, $k \in \{0, 1, 2, 3\}$ represents the sampled latency stage, and $i$ indexes the action horizon steps. $\tilde{s}_{k \cdot N}$ and $\mathcal{E}_0$ denote the time-misaligned proprioception and frozen semantic intent, respectively.
The micro-loop executes only the initial chunk of length $N=4$ before the next update. Therefore, we apply a horizon-aware mask $w_i$, setting $w_i = 2.0$ for $i < N$ and $1.0$ otherwise. This asymmetry forces the vector field to prioritize precision in the imminent execution window ($i < N$) over distal trajectory smoothness.

Standard Flow Matching samples the flow time step $\tau$ uniformly ($\tau \sim \mathcal{U}[0, 1]$), allocating equal capacity across the probability path. However, our micro-loop relies exclusively on single-step Euler integration starting from pure noise ($\tau=0$). The policy is only queried at the source frontier. Training capacity spent minimizing error at intermediate flow times (e.g., $\tau=0.5$) is therefore wasted, as these states are never visited during inference. To align the objective with our architecture, we bias the sampling distribution $p(\tau)$ heavily towards the noise source. We sample an auxiliary variable $u \sim \text{Beta}(\alpha, \beta)$ and set $\tau = 1 - u$. With $\alpha=5.0$ and $\beta=1.0$, the probability mass concentrates near $\tau \approx 0$. This ensures the vector field is most accurate exactly where the single-step solver queries it, maximizing the efficiency of the limited inference budget.

\subsection{Dynamic Perception via Motion Injection}
\label{sec:dynamic_perception}

We introduce a lightweight Differential Motion Predictor ($\phi_{\text{motion}}$) to inject high-frequency kinematic features directly into the policy.

We extract motion cues using temporal difference tensors. The input $\Delta I_t = \mathcal{T}(I_t) - \mathcal{T}(I_{t-\delta})$ is the normalized pixel-wise difference between the current frame and a history frame with lag $\delta=N$ (i.e., one micro-loop update interval). A low-latency CNN processes this input into a low-dimensional embedding $m_t \in \mathbb{R}^{64}$. We pretrain $\phi_{\text{motion}}$ using a loss to predict the target's future position and velocity, forcing $m_t$ to encode predictive physical dynamics and momentum.

To prevent control instability during manipulation where relative motion is zero, we employ a hard contact-gating mechanism. Based on a binary contact state $c_t \in \{0, 1\}$ derived from gripper sensors and the robot's proprioceptive joint state $s_{\text{prop}}$, we construct the fused policy state $\tilde{s}_t$ via concatenation:
\begin{equation}
    \tilde{s}_t = \text{Concat}\Big(s_{\text{prop}}, \; (1 - c_t) \cdot m_t\Big)
\end{equation}
This formulation enforces a modal switch: the policy attends to the full motion vector during the approach phase ($c_t=0$) and receives a zero-masked vector during manipulation ($c_t=1$).

\section{Experiments}

We evaluate whether TIDAL's decoupling and scheduling enable robust dynamic manipulation without losing generalist capabilities.

\subsection{Experimental Setup}

\subsubsection{Environments}
We evaluate TIDAL on two benchmarks in the RoboCasa suite.

We use 8 long-horizon tasks from the official \texttt{RoboCasa-GR1} benchmark \cite{Nasiriany-RSS-24, bjorck2025gr00t} with stationary targets to verify that our architecture retains generalist capabilities.

We construct a Dynamic Interception benchmark in which success requires the robot to intercept and grasp a moving target, place it into a drawer, and close the drawer. The target starts with a random cardinal-direction velocity and makes random 90-degree turns at boundaries, yielding non-deterministic trajectories. We stratify the benchmark into Easy and Hard tiers; using the left hand makes right/up motions kinematically favorable.

\subsubsection{Data Collection}
For the dynamic task, we collect 2,000 successful fine-tuning episodes using a paused-simulation oracle that pauses physics at every step to run full VLA inference, producing ideal closed-loop trajectories—a behavior achievable only when computational latency is ignored.

\subsubsection{Baselines and Computational Protocol}
We use GR00T-N1.5-3B \cite{bjorck2025gr00t} as the shared VLA backbone, fixing the prediction horizon at $H=16$ to isolate performance gains derived purely from scheduling efficiency. To quantify architectural latency, we use the latency numbers reported for NVIDIA Jetson AGX Orin in \cite{bjorck2025gr00t}. The Effective Update Frequency is derived from the total cycle time ($T_{\text{cycle}} = T_{\text{inference}} + T_{\text{execution}}$) at a 50\,Hz physical control rate (20\,ms per execution step). 

We compare TIDAL against three primary configurations:
\begin{itemize}
    \item \textbf{Open-Loop Baseline \cite{bjorck2025gr00t}:} Queries the VLM and a 4-step DiT of GR00T-N1.5-3B once at $t=0$, generating a 16-step chunk that is executed completely open-loop ($\sim$2.4\,Hz).
    \item \textbf{Compute-Heavy Replan:} A short-execution full-replanning variant that queries the full VLM and 4-step DiT every 4 physical steps. It generates a 16-step chunk but executes only the first 4 steps. It includes our motion predictor for dynamic tracking.
    \item \textbf{TIDAL (Ours):} Queries the VLM once every 16 steps to cache intent. Every 4 steps, it queries the flow head for a single Euler integration, updating the trajectory based on the latest observation.
\end{itemize}

\begin{table}[ht]
\centering
\caption{Inference Latency and Real-Time Update Profiles}
\label{tab:compute_cost}
\setlength{\tabcolsep}{4.5pt}
\begin{tabular}{l c c c c c}
\toprule
\textbf{Method} & \textbf{Cost} & \textbf{$T_{\text{inf}}$} (ms) & \textbf{$T_{\text{exec}}$} & \textbf{$T_{\text{cycle}}$} & \textbf{Update} \\ 
\midrule
Baseline & 1.0$\times$ & 93 & 320 & 413 & $\sim$2.4\,Hz \\
Comp-Heavy & $>$4.0$\times$ & 93 & 80 & 173 & $\sim$5.8\,Hz \\
\midrule
\textbf{TIDAL (Avg)} & \multirow{2}{*}{$\sim$1.05$\times$} & \textbf{$\sim$30} & \multirow{2}{*}{80} & \textbf{$\sim$110} & \textbf{$\sim$9.0\,Hz} \\
\textbf{TIDAL (Micro)} & & \textbf{$\sim$19} & & \textbf{$\sim$99} & \textbf{$\sim$10.1\,Hz} \\
\bottomrule
\end{tabular}
\end{table}

Table \ref{tab:compute_cost} shows that the Open-Loop Baseline has $T_{\text{cycle}}{=}413$\,ms, while Compute-Heavy reduces it to 173\,ms at an unsustainable $>4.0\times$ cost. TIDAL remains approximately iso-compute by amortizing one macro-step and four micro-steps, with $T_{\text{inf}}\!\approx\!30$\,ms on average (down to $\sim$19\,ms for micro-steps), enabling an effective $\sim$9--10\,Hz update rate without parallel hardware.

\subsection{Main Results}

Table \ref{tab:main_results} reports success rates on RoboCasa static tasks and our Dynamic Interception benchmark. To evaluate the true impact of inference latency, we report dynamic performance under two physics protocols: the idealized \textit{Paused} simulation where the environment waits for computation and the realistic \textit{Non-Paused} evaluation where physics advances during policy inference, approximating real-world deployment.

\begin{table*}[t]
\vspace{8pt}
\centering
\caption{Success Rate Comparison on Static and Dynamic Tasks under Different Physics Protocols}
\label{tab:main_results}
\definecolor{crashred}{HTML}{D62728}
\definecolor{robustblue}{HTML}{1F77B4}
\definecolor{lightgraybg}{HTML}{F2F2F2}

\begin{minipage}{0.95\linewidth}
\resizebox{\linewidth}{!}{
\begin{tabular}{l c c >{\columncolor{white}}c >{\columncolor{lightgraybg}}c c}
\toprule
\multirow{2}{*}{\textbf{Method}} & \textbf{Compute} & \textbf{Static Benchmark} & \multicolumn{2}{c}{\textbf{Dynamic (Easy)}} & \textbf{Dynamic (Hard)} \\ 
\cmidrule(lr){3-3} \cmidrule(lr){4-5} \cmidrule(lr){6-6}
& \textbf{Cost} & Standard & \textbf{Paused} \textit{(Ideal)} & \textbf{Non-Paused} \textit{(Realistic)} & \textbf{Paused} \textit{(Ideal)} \\ 
\midrule

Open-Loop Baseline & 1.0$\times$ & \textbf{0.6088} 
& 0.31 & 0.09 \textcolor{crashred}{\scriptsize ($\downarrow$71.0\%)} & 0.16 \\

Compute-Heavy Replan & \textcolor{crashred}{\textbf{$>$4.0$\times$}} & - 
& \textbf{0.63} & 0.17 \textcolor{crashred}{\scriptsize ($\downarrow$73.0\%)} & - \\

\midrule

\textbf{TIDAL (Ours)} & $\sim$1.05$\times$ & \underline{0.5938} \textcolor{gray}{\scriptsize ($\downarrow$2.5\%)}
& \underline{0.61} & \textbf{0.30} \textcolor{robustblue}{\scriptsize ($\downarrow$50.8\%)} & \textbf{0.36} \\

\bottomrule
\end{tabular}
}
\par\vspace{1.5ex}
\raggedright
{\scriptsize \textit{*Note:} \textbf{Bold}/\underline{underline} indicate best/second-best. The \textcolor{crashred}{\textbf{$>$4.0$\times$}} compute cost is highlighted as a severe penalty: while brute-force computation performs well in idealized \textit{Paused} physics, its inference latency leads to a severe \textcolor{crashred}{$\sim$73\%} drop in the realistic \colorbox{lightgraybg}{\textbf{Non-Paused}} environment. \textbf{TIDAL} robustly maintains performance with minimal compute overhead.\par}
\end{minipage}
\end{table*}

\subsubsection{Static Generalization}
On the official RoboCasa static benchmark, the Baseline achieves a 60.88\% success rate. TIDAL achieves 59.38\%, demonstrating that our hierarchical decoupling incurs only a slight drop. The slight drop is expected, as our temporally misaligned training objective and single-step flow integration create a more complex optimization landscape compared to standard open-loop training. However, this negligible trade-off confirms that TIDAL successfully retains the rich semantic priors of the 3B VLA backbone while freeing up computational budget for high-frequency control.

\subsubsection{Dynamic Interception (Idealized Paused Protocol)}
In the standard paused simulation, the Open-Loop Baseline fails in dynamic tasks because it blindly follows an outdated trajectory. To establish an upper performance bound, we introduce Compute-Heavy Replan, which uses aggressive full replanning—querying the full VLM and a 4-step DiT every 4 steps at a $>4.0\times$ compute cost—achieving a 0.63 success rate. TIDAL achieves an almost identical 0.61 at $\sim$1.05$\times$ compute cost. In a paused world, these results suggest that locally updating the flow field can approach the performance of full semantic replanning.

\subsubsection{Dynamic Interception (Realistic Non-Paused Protocol)}
The limitation of compute-heavy replanning is exposed under the non-paused protocol, where the environment continues to evolve during inference. The Baseline collapses to a 9\% success rate. The compute-heavy replan also experiences a severe drop from 0.63 to 0.17; although it updates frequently and perceives motion, its massive computational latency means targets move before new plans execute. Simply increasing replanning frequency without solving inference lag fails in physical reality. In contrast, TIDAL sustains a highly robust 0.30 success rate. By isolating high-frequency updates to the lightweight micro-loop, TIDAL minimizes reaction delay, actively correcting positional drift while the cached semantic plan remains valid.

\subsubsection{Failure Modes}
While interleaving substantially improves reactivity, the task remains a long-horizon manipulation problem and does not saturate even in a static setting (success $\sim$0.8) due to semantic and appearance variability. In the paused dynamic setting, remaining failures typically happen after interception during transport and placement into the right-side drawer. The object occasionally slips or is released, or ends up with a pose error that prevents a clean insert, consistent with suboptimal contact under dynamic grasps and limited grasp stability. Under the non-paused protocol, additional failures occur when the target moves too quickly or changes direction suddenly.

\subsection{Ablation Study}

We isolate the contributions of TIDAL's two core components: hierarchical inference scheduling and differential motion predictor. Table \ref{tab:ablations} shows the results on the Dynamic (Easy) task.

The TIDAL scheduling framework without motion features yields a success rate of 0.33, marginally better than the Baseline. Although operating at $\sim$9\,Hz, it lacks explicit velocity data. Injecting motion features into the Baseline improves performance to 0.43. While motion embeddings provide crucial initial velocity cues at $t=0$, open-loop execution prevents any mid-trajectory corrections. TIDAL achieves 0.61, outperforming both variants. This highlights a critical synergy: the motion predictor provides the signal to track dynamics, while the hierarchical architecture provides the computational opportunity to act on that signal reactively.

\begin{table}[ht]
\centering
\caption{Ablation Analysis on Dynamic (Easy) Tasks}
\label{tab:ablations}
\setlength{\tabcolsep}{4pt} 
\resizebox{\linewidth}{!}{
\begin{tabular}{l c c c l}
\toprule
\multirow{2}{*}{\textbf{Configuration}} & \textbf{Hierarchical} & \textbf{Motion} & \textbf{Success} & \multirow{2}{*}{\textbf{Key Limitation}} \\ 
 & \textbf{Micro-Loop} & \textbf{Predictor} & \textbf{Rate} & \\
\midrule
Baseline & \xmark & \xmark & 0.31 & High Latency + Static Features \\
w/o Motion & \cmark & \xmark & 0.33 & Motion Blindness \\
Base + Motion & \xmark & \cmark & 0.43 & Execution Blind Spot \\
\midrule
\textbf{TIDAL (Full)} & \cmark & \cmark & \textbf{0.61} & - \\
\bottomrule
\end{tabular}
}
\end{table}
 
\subsection{Semantic Intent Lifespan and Efficiency}

We analyze cached intent validity on a single static task from the official suite by increasing the number of physical steps executed per macro intent refresh. This forces the micro-loop to act longer under stale semantics, testing the policy's tolerance to intent staleness.

\begin{figure}[t]
\vspace*{4pt}
\centering
\includegraphics[width=\linewidth]{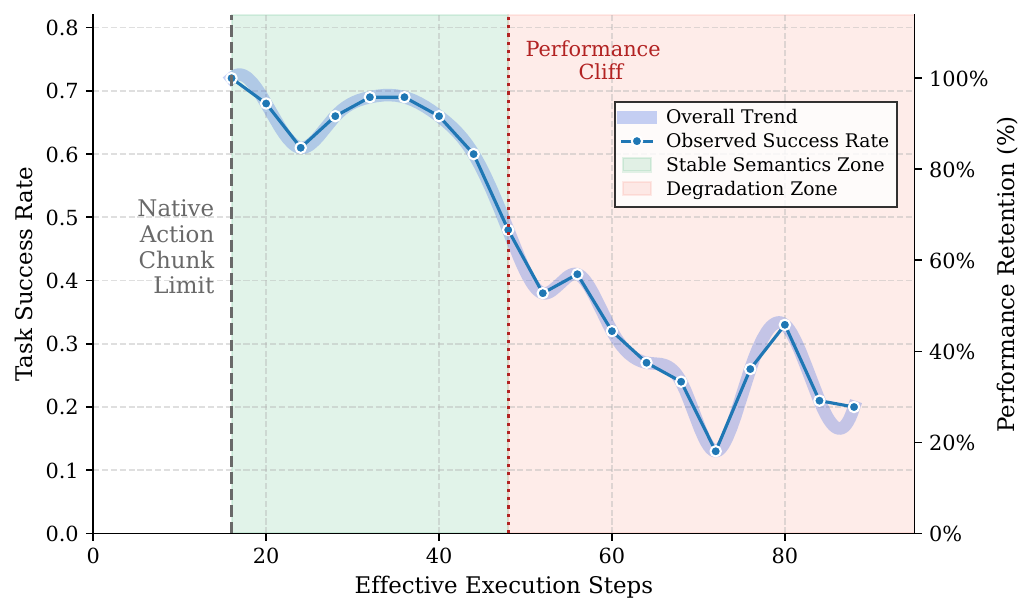}
\caption{Analysis of Semantic Intent Lifespan. The shaded regions indicate the Stable Semantics Zone (green) where intent remains valid, and the Degradation Zone (red) where performance collapses. The native action chunk limit is 16 steps.}
\label{fig:latency_efficiency}
\end{figure}

Fig. \ref{fig:latency_efficiency} shows a stable profile: extending intent lifespan from the standard 16 steps (0.72) to 44 steps (0.60) causes limited degradation. This suggests that cached embeddings remain useful when paired with real-time proprioceptive updates.

This finding challenges the native action chunk limit. While the GR00T backbone is trained to predict a fixed 16-step chunk, TIDAL's hierarchical architecture extends the useful horizon of a single intent query to $\sim$44 steps. The policy uses the old intent to condition new, locally correct flows. A sharp performance cliff appears at 48 steps, indicating where the robot's state diverges too far from the initial snapshot. Although variance increases at extreme latencies, the trend confirms that control ability degrades as semantic lag exceeds the physical correlation window.

\subsection{Real-World Deployment}

To validate interleaved scheduling under unmodeled real-world stochasticity, we deploy TIDAL on a KUKA iiwa 14 with a Robotiq 2F-140 gripper and an RGB camera.

\subsubsection{Task Setup and Environmental Stochasticity}
\begin{figure}[t]
\centering
\includegraphics[width=\linewidth]{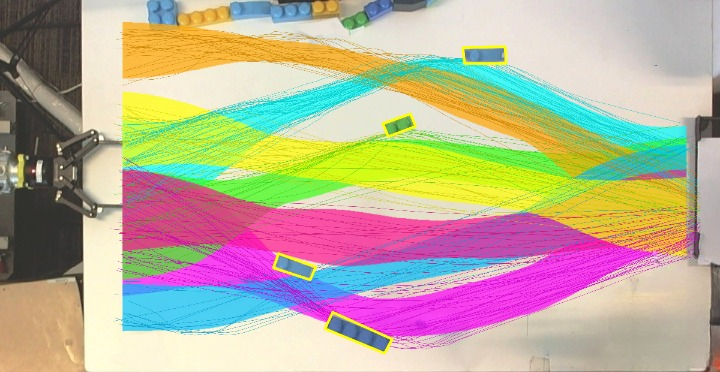}
\caption{\textit{Real-World Dynamic Interception Workspace.} The overhead view illustrates the stochastic nature of the task. Colored overlays represent diverse historical trajectories captured during evaluation, highlighting unmodeled non-linearities, decelerations due to friction, and sharp deflections caused by block obstacles.}
\label{fig:real_workspace}
\end{figure}

As shown in Fig. \ref{fig:real_workspace}, objects roll down a ramp into a flat workspace with block obstacles, introducing trajectory uncertainty from friction, spin, and obstacle deflections. To isolate scheduling effects, we also equip the open-loop baseline with the motion predictor. It still executes the full 16-step chunk, whereas TIDAL executes only $N=4$ steps before each micro-loop update.

\begin{figure}[t]
\vspace*{4pt}
\centering
\includegraphics[width=\linewidth]{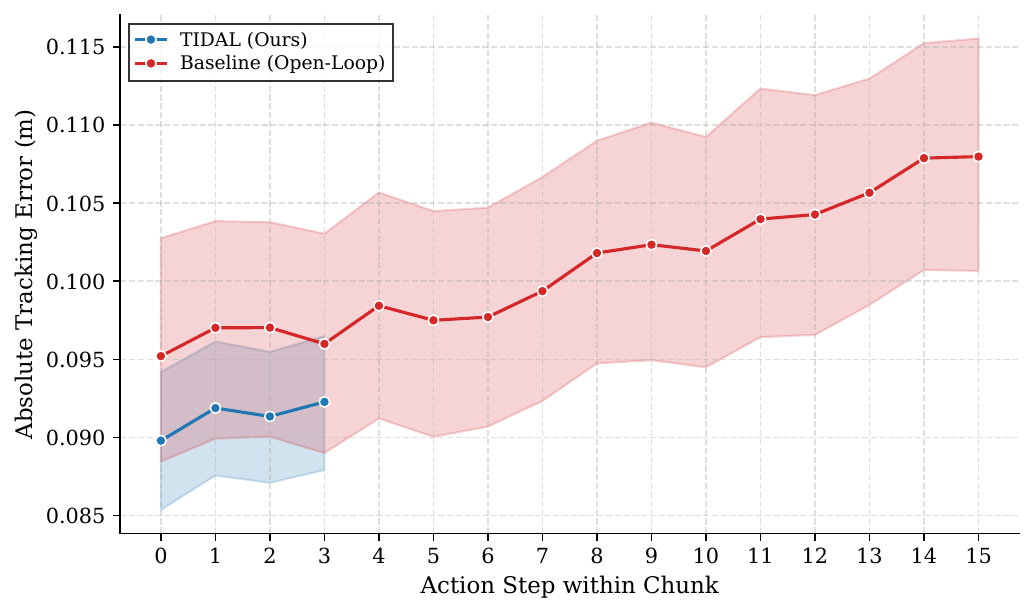}
\caption{Temporal Degradation of Action Chunks. Tracking error vs. action index diverges for the open-loop 16-step rollout, while TIDAL executes only the first 4 steps to bound error within gripper tolerance.}
\label{fig:chunk_degradation}
\end{figure}

\subsubsection{Reactivity and Error Bounding}
In real-world experiments, TIDAL achieves an 86\% success rate, substantially outperforming the motion-augmented open-loop baseline (34\%) over 100 trials, yielding roughly a 2.5$\times$ improvement in dynamic environments.

The motion-augmented baseline is accurate under near-linear rolls but fails once real perturbations (friction/spin/collisions) change the target trajectory. Executing a 16-step chunk open-loop, it accumulates tracking error throughout the rollout (Fig. \ref{fig:chunk_degradation}). Once the target deviates from its initial motion at $t=0$, the later chunk steps effectively become stale open-loop extrapolation.

TIDAL instead bounds error by truncating execution to $N{=}4$ steps and triggering a micro-loop update (Fig. \ref{fig:chunk_degradation}). Although tracking error can increase slightly within each 4-step window, the next update reconditions on fresh proprioception and motion, preventing divergence. This frequent refresh keeps error within the gripper’s tolerance under unmodeled disturbances.

\section{Conclusion}
\balance

We presented TIDAL, a framework designed to enable high-frequency control with large VLA models. By interleaving single-step flow generation with action execution, our approach achieves a 4x increase in update frequency on edge hardware while maintaining the same backbone computational budget. This significantly reduces the execution blind spot found in standard batch-and-execute baselines.

We leverage temporally misaligned training to compensate for stale semantics, while single-step flow integration minimizes latency. This results in a 2.5x performance gain on dynamic interception tasks. Our analysis reveals that updating the policy with real-time proprioception effectively extends the lifespan of cached semantic embeddings beyond their native horizon. Since TIDAL is backbone-agnostic, it remains orthogonal to system-level accelerations. TIDAL sustains robust performance under non-paused inference protocols, where compute-heavy baselines degrade sharply due to latency. Furthermore, real-world deployment on a physical manipulator confirms these findings. By caching temporally persistent semantic intent and interleaving low-latency flow updates with execution, TIDAL turns long-horizon VLA plans into a high-rate closed-loop controller, keeping tracking error bounded under non-paused inference and unmodeled real-world dynamics without increasing backbone queries.

\section*{ACKNOWLEDGMENT}
OpenAI's ChatGPT was used to assist with implementation debugging and English-language editing. All technical content, analyses, and conclusions were reviewed and verified by the authors. This research is partly supported by the National Research Foundation, Singapore, under its Thematic CRP programme award no NRF-T-CRP-2025-0008 and by A*STAR under the National Robotics Programme grant no H25-MNR3495.

\bibliographystyle{IEEEtran}
\bibliography{references} 

\end{document}